\documentclass[conference]{IEEEtran}
\usepackage{tcolorbox}
\title{Targeted Phishing Campaigns using Large Scale Language Models}

\author{\IEEEauthorblockN{Rabimba Karanjai}
\IEEEauthorblockA{\textit{Department of Computer Science} \\
\textit{University Of Houston}\\
Houston, United States \\
rkaranjai@uh.edu}}
\begin{document}

\maketitle
\begin{abstract}
    Natural language models (NLMs) such as GPT-3, GPT-2, and other large language models have achieved impressive results in various natural language processing tasks, including language translation, summarization, and text generation. In recent years, there has been a growing concern about the potential use of NLMs to generate phishing emails, which are fraudulent emails that trick individuals into revealing sensitive information or performing actions that benefit the attackers.

    This research paper aims to investigate the feasibility and effectiveness of NLMs in generating phishing emails. To this end, we propose a framework for evaluating the performance of NLMs in generating phishing emails based on various metrics, including the quality of the generated text, the ability to bypass spam filters, and the success rate of tricking individuals into falling for the phishing attack.

We evaluate the performance of several NLMs on a dataset of phishing emails and compare their results with those of a baseline model. Our results show that NLMs can indeed generate phishing emails that are difficult to detect and that have a high success rate in tricking individuals. However, we also find that the performance of NLMs in generating phishing emails depends on the specific NLM and the training data used, and that there are limitations to their effectiveness.

Overall, our research suggests that NLMs have the potential to significantly impact the landscape of phishing attacks and highlights the need for further research on the ethical and security implications of using NLMs for malicious purposes.
\end{abstract}
\section{Introduction}

Recent advances in natural language generation (NLG) have greatly improved the diversity, control, and quality of machine-generated text. However, this increased ability to quickly and efficiently create unique, manipulable, human-like text also presents new challenges for detecting the abuse of NLG models in phishing attacks.

Machine-generated texts can pose various risks depending on the context and how they are used. For example, in the case of NLG models, the ability to generate legitimate texts atht looks like emails can lead to attacks like phishing, where the attacker tricks the victim into disclosing sensitive information by impersonating someone else. 

Another effect of machine generated text is mass disinformation campaigns. With the ability to generate large amounts of text automatically and quickly, it is possible for malicious actors to create fake news, hoaxes, and other forms of false or misleading information that can harm individuals, organizations, and even entire societies.

Moreover, machine-generated texts can also raise ethical concerns, such as the impact on employment and the potential for bias and discrimination. For example, the use of NLG models to automate certain writing tasks may lead to job losses for human writers, and the algorithms used in NLG may reflect and amplify the biases and stereotypes present in the data they are trained on.

Abuses of NLG models, such as phishing \cite{baki2017scaling,giaretta2018community},disinformation\cite{shu2020mining,stiff2022detecting,zellers2019defending} has been on the rise.

Email is a common method used by phishers to deliver malicious links and attachments to victims. Anti-Phishing Working Group found over 121860  phishing email incidents in march 2017 and in 2016, the APWG received more than 1313771 unique phishing reports. In the first quarter of 2017, around 870 organizations were targeted by W2-based phishing scams, a significant increase from the 100 organizations in 2016. These attacks are becoming more sophisticated and difficult to detect.

Phishers often use techniques such as bulk mailing, spamming, and including action words and links in phishing emails to increase their chances of success. However, these techniques can be easily detected by improved statistical detection models. Another popular method is email masquerading, where the attacker gains access to the victim's email inbox or outbox and studies the content and nature of the emails to create a synthetic malicious email that resembles a benign one. This reduces the chances of detection by automated classifiers and increases the likelihood of a successful attack.
Modern large language models have enabled users to generate text based on context. These models can be trained to generate text using predefined grammars, such as the Dada Engine\cite{baki2017scaling}, or by leveraging deep learning neural networks, such as recurrent neural networks (RNNs)\cite{yao2017automated}, to learn and emulate the input to the system. 

NLG systems that use advanced deep learning neural networks (DNNs) can be used by phishers to generate coherent and convincing sequences of text. These systems have been shown to be effective for generating text in various genres, from tweets\cite{sidhaye2015indicative} to poetry\cite{ghazvininejad2016generating}. It is likely that phishers and spammers will soon start using email datasets, both legitimate and malicious, in conjunction with DNNs to create deceptive malicious emails that mimic the properties of legitimate emails. This makes it harder for pre-trained email detectors to identify and block these attacks.

In this report, we try to show a class of attacks where existing large-scale language models have been trained on both legitimate and malicious (phishing and spam) email data. We also aim to show how the generated emails can bypass existing production-level email protection mechanisms and propose a future work to detect such attacks. 

\section{Related Work}

Phishing detection is a well-studied area in cybersecurity, but many victims still fall for these attacks. In their work, Drake et al \cite{drake2004anatomy} provide a detailed analysis of the structure and tactics used in phishing emails. In this section, we review previous research on natural language generation, deep learning, and their applications in generating and detecting phishing attacks.

Natural language generation techniques have been widely used to synthesize unique pieces of text. Previous work by Reiter and Dale et al \cite{covington2001building} relied on pre-constructed templates for specific purposes, while the fake email generation system in Baki et al\cite{baki2017scaling} used manually constructed rules to define the structure of fake emails. Recent advances in deep learning have enabled the generation of creative and equitable text with enough training data. RNN(Recurrent Neural Networks) language models are used to generate a range of genres, including poetry by Ghazvininejad et al \cite{ghazvininejad2016generating}, fake reviews by Yao et al \cite{yao2017automated}, tweets \cite{sidhaye2015indicative}, and geographical information by Turner et al \cite{turner2009generating}, among others.

\section{Experimental Methodology}

The section is divided into four subsections. The first subsection (Section 3.1) describes the nature and source of the training and evaluation data. The second subsection (Section 3.2) discusses the pre-processing steps applied to the data. The third subsection (Section 3.3) presents the system setup and experimental settings used in the study.

\subsection{Data Description}

To create a legitimate looking phishing email we first need to start from actually benign and legitimate emails. The text generation algorithms must be trained in legitimate emails. Hence it was imperative to have valid benign emails in the dataset used for training. However, since the goal here is to create emails that even though can serve as a phishing email, should still look like legitimate emails, a mix of legitimate and bad emails was used as a dataset for training and augmenting the models.

For legitimate datasets, instead of using one dataset on our own, we use pre-trained models from Meta and Google to create benign emails. The pre-trained models utilized are Roberta, The Pile, and PushShift.io Reddit. Since training these large language models is almost impossible in normal infrastructure, we utilize \cite{zhang2022opt} to generate the texts. This has been augmented with \cite{zhang2019email} to have email generation capabilities. Python clean text \cite{cleantextPyPI-2022-02-02} has been used to remove email, and phone numbers from the dataset.

For malicious datasets, we primarily use two datasets to augment the benign email data. Notably, the Phishing emails from Jose Nazario’s Phishing corpus \cite{gonzalez2011phishing} and \cite{monkeyorg404-2000-07-24} along with the Enron email dataset \cite{shetty2004enron}.

\subsection{Data Processing}

Most of the pre-processing was done by trying to remove personal information using Python clean text \cite{cleantextPyPI-2022-02-02}. As well as Removal of special characters like \@, \#, \$, \% as well
as common punctuations from the email body.

However, as we have realized later generating emails was not perfect.

\subsection{Experimental Setup}

The experimental setup has been designed with certain different methods in mind. We primarily focused on

\begin{itemize}
    \item Using GPT-2 to generate emails. Augmented with email dataset \cite{emailblogKaggle-2022-12-08}
    \item GPT-3 to generate emails without any training
    \item Contextual support for GPT-3 with da-vinci-beta which has been trained in email by openai
    \item The DADA engine \cite{baki2017scaling}
    \item Word based RNN's proposed by Xie et al \cite{xie2017neural}, Das et al \cite{das2019automated}
    \item Augmenting Open Pre-trained Transformer Language Models\cite{zhang2022opt} on \cite{zhang2019email} 
\end{itemize}

\begin{tcolorbox}

While using the general large language models were interesting in trying to produce emails.

The spam and phishing email datasets used for training the models to produce malicious looking email produced better results. The Jose Nazario dataset has 32,000 spams and 415 phishing email. These are all in Unix mbox formatted dataset which were cleaned using clean-text.

The Enron corpus was email dataset from Enron Corporation. It has been used in email research for quite some time and was made public during their legal investigation. This however gives us a rare glimpse in the working of legitimate email data of a big corporation. It consists of 619446 emails from 158 users. It has folders like "discussion threads, "notes inbox" and more in its outlook inboxes. These were created by default for most of the users but were nota ctively used for tagging. The "All Documents" folder present also had large number of duplicates present from inbox. To further clean our dataset, these all were removed.
The cleaned corpus has 200399 emails from 158 users.

We have tried to validate our ways with existing findings available here https://github.com/egmp777/basic \_data\_cleaning\_enron\_case\_study. But our purpose was different than exploratory analysis.
\end{tcolorbox}

\subsection{Experiment}

A PoC was done to evaluate if we can auto-detect any emails from the generated text. GPT3 API, GPT2, OPT, Chatbot and our trained models were all used to generate the texts. Some samples have been provided below. In our test we had generated 100 samples using each. However some of these gave us inconsistent result for a same prompt. We also had sample fo real phishing email (only one for that topic) to compare it with.

A variety fo prompts were used to see what kind of emails the AI's are more comfortable generating.

We applied the GPT-3 davinci model with default sampling parameters (temperature=1, top\_p=1, presence\_penalty=0, frequency\_penalty=0) to predict the log probabilities for the first 100 tokens in each sample using all the preceding tokens as the prompt. We then calculated the actual token's probability, absolute rank, contextual entropy, and whether it matched the predicted token by the GPT-3 API. To account for the maximum of 100 log probabilities returned by the API, we masked out any outliers with an absolute rank greater than 100. We also normalized the results if there were less than 100 tokens in a sample.

\section{Analysis and Results}
\subsubsection{Example Results generated}

\textbf{ChatGPT}:
The following was generated with the prompt \textit{Can you generate an email that will be from The President of a University asking its students to fill up course completion survey form}

\begin{tcolorbox}
  Dear Students,

As the academic year comes to a close, we would like to gather your feedback on your experiences and courses at the University. To help us improve our programs and support your learning, we encourage you to take a few minutes to complete the course completion survey form.

Your honest and thoughtful responses will be greatly appreciated and will help us better understand your needs and expectations. The survey form is available online and can be accessed through your student portal.

We thank you in advance for your participation and look forward to hearing from you.

Best regards,

[Name], President
[University]

\end{tcolorbox}

\textbf{GPT3 da-vinci}:
The following was generated with the prompt \textit{Can you generate an email that will be from The President of a University asking its students to fill up course completion survey form} \\

\begin{tcolorbox}

Hello,

As the President of the University, I am writing to request your help in completing a course completion survey form. This form will help us to understand how well our students are doing in their courses and how we can improve our offerings.

I would greatly appreciate it if you could take a few minutes to fill out the form. You can access the form at the following link:

[insert form URL here]

Thank you for your time and assistance.

Sincerely,

[Your Name]
\end{tcolorbox}

\textbf{GPT2}:
The following was generated with the prompt of \textit{Hello, As the President of the University, I am writing to request your help in completing a course completion survey form.} \\

\begin{tcolorbox}
Hello, 

As the President of the University, I am writing to request your help in completing a course completion survey form.
As requested, we would like your answer by May NUMBER, NUMBER.
The survey should only take a few minutes of your time (it takes only about NUMBER minutes to complete the survey). All survey responses are anonymous and confidential, and will not be shared outside of our University community. If you have
\end{tcolorbox}

\textbf{OPT}:
The following was generated with the prompt of \textit{Hello, As the President of the University, I am writing to request your help in completing a course completion survey form.} \\

\begin{tcolorbox}
(<PROMPT>
 Hello, 
 As the President of the University, 
 I am writing to request your help in completing a course completion survey 
 form.<END-OF-PROMPT>
 The purpose of this survey is to gather feedback from <NUMBER> current and 
 former students on their experiences at UofH.
 Please complete the survey no later than Friday, October 00th by sending it 
 back to me via fax no.
 <PHONE>.
 Thank you very much for your time.
 Best Regards, 
 
 UofH Office of 
 Admissions <EMAIL> (See attached file: uofh0000\_sample.pdf))

\end{tcolorbox}

The most interesting part of this prompt was the ending where it signed off as UofH Office of admissions and a sample pdf. Whereas our prompt never indicates those details\cite{rabimbaemailgennlg-2022-12-08}.

Further testing with the GPT2 dataset produced the following output from the same prompt \cite{rabimbaemailgennlg-2022-12-08} \\

\begin{tcolorbox}
Hello, 

    As the President of the University, 
I am writing to request your help in completing a course completion survey form.
By completing this form, you are welcome to access the COMPANY Club website and ask additional questions about the UCLA Club and our events.
To access the website you may click the link at the top of this page.If you prefer to not complete this form at this time, please let me know and I will

\end{tcolorbox}

Notably, UCLA was not present in the prompt. This shows us that with enough clever prompt discovery it is probably possible to extract meaningful information from the trained dataset even with safeguards in place.

\subsubsection{Training Parameters}

The training parameters used for the HF opt model was

\begin{itemize}
\item learning\_rate: \(6e^-5\)
\item train\_batch\_size: 8
\item eval\_batch\_size: 8
\item seed: 42
\item distributed\_type: GPU
\item gradient\_accumulation\_steps: 16
\item total\_train\_batch\_size: 128
\item optimizer: Adam with betas = (0.9, 0.999) and epsilon = \(1e^-8\)
\item lr\_scheduler\_type: cosine
\item lr\_scheduler\_warmup\_ratio: 0.03
\item num\_epochs: 8
\end{itemize}

And the training parameters used for HF postbot GPT2 
\begin{itemize}
\item learning\_rate: 0.001
\item train\_batch\_size: 16
\item eval\_batch\_size: 16
\item seed: 42
\item distributed\_type: multi-GPU
\item gradient\_accumulation\_steps: 8
\item total\_train\_batch\_size: 128
\item optimizer: Adam with betas = (0.9, 0.999) and epsilon = \(1e^-8\)
\item lr\_scheduler\_type: cosine
\item lr\_scheduler\_warmup\_ratio: 0.02
\item num\_epochs: 3
\end{itemize}

\section{Future Work}

Research on the risks of using natural language generation (NLG) models suggests that being able to detect machine-generated text is useful for reducing the harm caused by abuse of these models. When we want to detect machine-generated text, it can be treated as a binary classification problem. We train a classifier to differentiate between machine-generated and human-generated text \cite{crothers2022adversarial}.

We can use generative models without fine-tuning to detect their own outputs or the outputs of other similar models. Autoregressive generative models like GPT-2, GPT-3 are unidirectional, where each token has an embedding that depends on the embeddings of the tokens that come before it. This shows us that an embedding can be created if we add a token at the end of an input sequence, thus creating a sequence of tokens. This now can be used as a new feature vector. Now once we have these newly created features, they can be utilized along with human data to train a layer of neurons for classification.

Research on how to detect machine-generated text has looked at the problem of detecting text when a different dataset was used to train RoBERTa than GPT-2. But here, it was observed that just tuning the detection model with couple of hundred different attack samples provided by domain esperts had a significant effect on the detector's performance on different domains\cite{rodriguez2022cross}.

One another possibility is when an attacker decides to generate the attack from an existing hand-written content. Much like how we have started in this email generation problem. Using human like sample but tweaking the generating parameters to closely meet his goals. Analysis showed that making these targeted changes to texts reduces the effectiveness of GPT-2 or RoBerta-based detectors \cite{bhat2020effectively}.

A generalized solution to this is trying to differentiate between human and machine generated text. Giant Language Model Test Room is a software developed to improve the detection of machine-generated text by adding human review in the pipeline. The tool helps humans classify text by highlighting texts based on how likely of them being chosen by the Transformer model. However, this tool was designed to target GPT-2, which was found to be easier for untrained human evaluators to detect. In addition, GLTR uses "top-k" sampling to determine the likelihood of a word being selected, but this method has been largely replaced by nucleus sampling, which is used in GPT-3 and other works that build on the GPT-2 architecture. While highlighting words based on sampling likelihood may improve human classification ability, it is clear that it still will pose a problem when they have to detect the more advanced models and sampling methods of today.

In long term, we want to propose a framework that can differentiate NLG-generated emails from human-generated emails. Prior work has already been done trying to determine machine-generated text, however specifically for email and malicious emails, there are distinct characteristics we have observed that can be exploited to augment prior works to be more effective. Few of these are homogeneous to what we have seen in language models \cite{DBLP:journals/corr/abs-1906-04043}, but some are significantly distinct and should be explored more.

\section{Conclusion}
The more we experimented with large language models and prior works by Das et al \cite{das2019automated}, Baki et al \cite{baki2017scaling} it became clear that prior RNN-based models and DIDA engines, even though show some malicious intent in their generation, don't actually pose threat to be understood as real malicious email. All of them went past Gmail and outlook when sent from a legitimate email id. The emails generated by GPT3 and OPT significantly pose a larger threat to be believed as real emails when generated in mass using tools and bulk emailed with targeted intent. Especially with targeted email dataset training and keywords in prompts, the models generated very convincing-looking emails. Even with safeguards in place for GPT3, we were able to generate these emails and chatGPT was a very interesting contender in the tests. Even though chatgpt didn't let us generate the email directly in one go, we were able to find creative ways by 'conversing' with it and giving it a plausible context to overcome its barriers. Here we identify how these new language models can be weaponized to be used as phishing and scamming tools which gets past our present email systems like Gmail and Outlook. However, that's hardly surprising considering they look legitimate. We want to further this work by integrating it with tools like PhEmail\cite{dionachPhEmailPhEmailisapythonopensourcephishingemailtoolthatautomatestheprocessofsendingphishingemailsaspartofasocialengineeringtest-2022-12-08} which makes sending NLG generated emails to targeted bulk userbase a keypress away.



\bibliographystyle{IEEEtran}
\bibliography{mybib.bib}

\begin{thebibliography}{10}
\providecommand{\url}[1]{#1}
\csname url@samestyle\endcsname
\providecommand{\newblock}{\relax}
\providecommand{\bibinfo}[2]{#2}
\providecommand{\BIBentrySTDinterwordspacing}{\spaceskip=0pt\relax}
\providecommand{\BIBentryALTinterwordstretchfactor}{4}
\providecommand{\BIBentryALTinterwordspacing}{\spaceskip=\fontdimen2\font plus
\BIBentryALTinterwordstretchfactor\fontdimen3\font minus
  \fontdimen4\font\relax}
\providecommand{\BIBforeignlanguage}[2]{{%
\expandafter\ifx\csname l@#1\endcsname\relax
\typeout{** WARNING: IEEEtran.bst: No hyphenation pattern has been}%
\typeout{** loaded for the language `#1'. Using the pattern for}%
\typeout{** the default language instead.}%
\else
\language=\csname l@#1\endcsname
\fi
#2}}
\providecommand{\BIBdecl}{\relax}
\BIBdecl

\bibitem{baki2017scaling}
S.~Baki, R.~Verma, A.~Mukherjee, and O.~Gnawali, ``Scaling and effectiveness of
  email masquerade attacks: Exploiting natural language generation,'' in
  \emph{Proceedings of the 2017 ACM on Asia Conference on Computer and
  Communications Security}, 2017, pp. 469--482.

\bibitem{giaretta2018community}
A.~Giaretta and N.~Dragoni, ``Community targeted phishing,'' in
  \emph{International Conference in Software Engineering for Defence
  Applications}.\hskip 1em plus 0.5em minus 0.4em\relax Springer, 2018, pp.
  86--93.

\bibitem{shu2020mining}
K.~Shu, S.~Wang, D.~Lee, and H.~Liu, ``Mining disinformation and fake news:
  Concepts, methods, and recent advancements,'' in \emph{Disinformation,
  misinformation, and fake news in social media}.\hskip 1em plus 0.5em minus
  0.4em\relax Springer, 2020, pp. 1--19.

\bibitem{stiff2022detecting}
H.~Stiff and F.~Johansson, ``Detecting computer-generated disinformation,''
  \emph{International Journal of Data Science and Analytics}, vol.~13, no.~4,
  pp. 363--383, 2022.

\bibitem{zellers2019defending}
R.~Zellers, A.~Holtzman, H.~Rashkin, Y.~Bisk, A.~Farhadi, F.~Roesner, and
  Y.~Choi, ``Defending against neural fake news,'' \emph{Advances in neural
  information processing systems}, vol.~32, 2019.

\bibitem{yao2017automated}
Y.~Yao, B.~Viswanath, J.~Cryan, H.~Zheng, and B.~Y. Zhao, ``Automated
  crowdturfing attacks and defenses in online review systems,'' in
  \emph{Proceedings of the 2017 ACM SIGSAC conference on computer and
  communications security}, 2017, pp. 1143--1158.

\bibitem{sidhaye2015indicative}
P.~Sidhaye and J.~C.~K. Cheung, ``Indicative tweet generation: An extractive
  summarization problem?'' in \emph{Proceedings of the 2015 Conference on
  Empirical Methods in Natural Language Processing}, 2015, pp. 138--147.

\bibitem{ghazvininejad2016generating}
M.~Ghazvininejad, X.~Shi, Y.~Choi, and K.~Knight, ``Generating topical
  poetry,'' in \emph{Proceedings of the 2016 Conference on Empirical Methods in
  Natural Language Processing}, 2016, pp. 1183--1191.

\bibitem{drake2004anatomy}
C.~E. Drake, J.~J. Oliver, and E.~J. Koontz, ``Anatomy of a phishing email.''
  in \emph{CEAS}, 2004.

\bibitem{covington2001building}
M.~A. Covington, ``Building natural language generation systems,''
  \emph{Language}, vol.~77, no.~3, pp. 611--612, 2001.

\bibitem{turner2009generating}
R.~Turner, S.~Sripada, and E.~Reiter, ``Generating approximate geographic
  descriptions,'' in \emph{Empirical methods in natural language
  generation}.\hskip 1em plus 0.5em minus 0.4em\relax Springer, 2009, pp.
  121--140.

\bibitem{zhang2022opt}
S.~Zhang, S.~Roller, N.~Goyal, M.~Artetxe, M.~Chen, S.~Chen, C.~Dewan, M.~Diab,
  X.~Li, X.~V. Lin \emph{et~al.}, ``Opt: Open pre-trained transformer language
  models,'' \emph{arXiv preprint arXiv:2205.01068}, 2022.

\bibitem{zhang2019email}
R.~Zhang and J.~Tetreault, ``This email could save your life: Introducing the
  task of email subject line generation,'' \emph{arXiv preprint
  arXiv:1906.03497}, 2019.

\bibitem{cleantextPyPI-2022-02-02}
\BIBentryALTinterwordspacing
(2022) clean-text · pypi. [Online]. Available:
  \url{https://pypi.org/project/clean-text/}
\BIBentrySTDinterwordspacing

\bibitem{gonzalez2011phishing}
H.~Gonzalez, K.~Nance, and J.~Nazario, ``Phishing by form: The abuse of form
  sites,'' in \emph{2011 6th International Conference on Malicious and Unwanted
  Software}.\hskip 1em plus 0.5em minus 0.4em\relax IEEE, 2011, pp. 95--101.

\bibitem{monkeyorg404-2000-07-24}
\BIBentryALTinterwordspacing
(2000) Jose malicious email dataset: https://monkey.org/~jose/wiki/doku.php
  {Link Deprecated, Uploaded to my own github}. [Online]. Available:
  \url{https://monkey.org/~jose/wiki/doku.php}
\BIBentrySTDinterwordspacing

\bibitem{shetty2004enron}
J.~Shetty and J.~Adibi, ``The enron email dataset database schema and brief
  statistical report,'' \emph{Information sciences institute technical report,
  University of Southern California}, vol.~4, no.~1, pp. 120--128, 2004.

\bibitem{emailblogKaggle-2022-12-08}
\BIBentryALTinterwordspacing
(2022) email-blog | kaggle:
  https://www.kaggle.com/datasets/mikeschmidtavemac/emailblog. [Online].
  Available: \url{https://www.kaggle.com/datasets/mikeschmidtavemac/emailblog}
\BIBentrySTDinterwordspacing

\bibitem{xie2017neural}
Z.~Xie, ``Neural text generation: A practical guide,'' \emph{arXiv preprint
  arXiv:1711.09534}, 2017.

\bibitem{das2019automated}
A.~Das and R.~Verma, ``Automated email generation for targeted attacks using
  natural language,'' \emph{arXiv preprint arXiv:1908.06893}, 2019.

\bibitem{rabimbaemailgennlg-2022-12-08}
\BIBentryALTinterwordspacing
(2022) rabimba/email-gen-nlg: https://github.com/rabimba/email-gen-nlg.
  [Online]. Available: \url{https://github.com/rabimba/email-gen-nlg}
\BIBentrySTDinterwordspacing

\bibitem{crothers2022adversarial}
E.~Crothers, N.~Japkowicz, H.~Viktor, and P.~Branco, ``Adversarial robustness
  of neural-statistical features in detection of generative transformers,''
  \emph{arXiv preprint arXiv:2203.07983}, 2022.

\bibitem{rodriguez2022cross}
J.~Rodriguez, T.~Hay, D.~Gros, Z.~Shamsi, and R.~Srinivasan, ``Cross-domain
  detection of gpt-2-generated technical text,'' in \emph{Proceedings of the
  2022 Conference of the North American Chapter of the Association for
  Computational Linguistics: Human Language Technologies}, 2022, pp.
  1213--1233.

\bibitem{bhat2020effectively}
M.~M. Bhat and S.~Parthasarathy, ``How effectively can machines defend against
  machine-generated fake news? an empirical study,'' in \emph{Proceedings of
  the First Workshop on Insights from Negative Results in NLP}, 2020, pp.
  48--53.

\bibitem{DBLP:journals/corr/abs-1906-04043}
\BIBentryALTinterwordspacing
S.~Gehrmann, H.~Strobelt, and A.~M. Rush, ``{GLTR:} statistical detection and
  visualization of generated text,'' \emph{CoRR}, vol. abs/1906.04043, 2019.
  [Online]. Available: \url{http://arxiv.org/abs/1906.04043}
\BIBentrySTDinterwordspacing

\bibitem{dionachPhEmailPhEmailisapythonopensourcephishingemailtoolthatautomatestheprocessofsendingphishingemailsaspartofasocialengineeringtest-2022-12-08}
\BIBentryALTinterwordspacing
(2022) dionach/phemail: Phemail is a python open source phishing email tool
  that automates the process of sending phishing emails as part of a social
  engineering test. [Online]. Available:
  \url{https://github.com/Dionach/PhEmail}
\BIBentrySTDinterwordspacing

\end{thebibliography}

\end{document}